# KuiSCIMA v2.0: Improved Baselines, Calibration, and Cross-Notation Generalization for Historical Chinese Music Notations in Jiang Kui's Baishidaoren Gequ

Tristan Repolusk[✉][1,2][0009−0009−8435−1185] and
Eduardo Veas[1,2][0000−0002−0356−4034]

[1] Graz University of Technology, Graz, Austria
`tristan.repolusk@student.tugraz.at, eveas@tugraz.at`
[2] Know Center Research GmbH, Graz, Austria
`{trepolusk,eveas}@know-center.at`

**Abstract.** Optical Music Recognition (OMR) for historical Chinese musical notations, such as *suzipu* 俗字谱 and *lülüpu* 律吕谱, presents unique challenges due to high class imbalance and limited training data. This paper introduces significant advancements in OMR for Jiang Kui's influential collection *Baishidaoren Gequ* 白石道人歌曲 from 1202. In this work, we develop and evaluate a character recognition model for scarce imbalanced data. We improve upon previous baselines by reducing the Character Error Rate (CER) from 10.4% to 7.1% for *suzipu*, despite working with 77 highly imbalanced classes, and achieve a remarkable CER of 0.9% for *lülüpu*. Our models outperform human transcribers, with an average human CER of 15.9% and a best-case CER of 7.6%. We employ temperature scaling to achieve a well-calibrated model with an Expected Calibration Error (ECE) below 0.0162. Using a leave-one-edition-out cross-validation approach, we ensure robust performance across five historical editions. Additionally, we extend the KuiSCIMA dataset to include all 109 pieces from *Baishidaoren Gequ*, encompassing *suzipu*, *lülüpu*, and *jianzipu* notations. Our findings advance the digitization and accessibility of historical Chinese music, promoting cultural diversity in OMR and expanding its applicability to underrepresented music traditions.

## 1 Introduction

Optical music recognition (OMR), a subfield of artificial intelligence, deals with computationally extracting symbolic musical notation in documents that are given as images. Its application includes the documentation of music sheets for storage or transcribing the music sheet to suitable symbolic formats [4]. However, the majority of the works concern common practice period music scores, and most do not support lesser-known or early forms of notations.

In this paper we focus on OMR methods acting on the poetry collection *Baishidaoren Gequ* first published in 1202. This is an important work in the history of Chinese music notation and composition compiling the works of Jiang Kui, a renowned poet, calligrapher, and music theorist of the Southern Song dynasty (1127-1279 CE). This collection is one of the earliest surviving examples of melodized lyrics in Chinese history [11]. It contains three different Chinese notations (*suzipu*, *lülüpu*, and *jianzipu* 减字谱) and has been published in various handwritten editions in the last few centuries. These notations feature complex spatial layouts, high class imbalance, and limited annotated data, making traditional OMR methods inadequate. In this paper, we address the challenges of recognizing historical Chinese notations, specifically *suzipu* and *lülüpu*, from Jiang Kui's *Baishidaoren Gequ* (1202).

We present improved OMR models that significantly reduce the baseline Character Error Rate (CER) for *suzipu* (from 10.4% to 7.1%) and achieve a CER of 0.9% for *lülüpu*. Our models outperform naive human transcribers, with an average human CER of 15.9% and a best-case CER of 7.6%. To address data scarcity, we employ lightweight CNN architectures, temperature scaling for model calibration, and a leave-one-edition-out cross-validation approach to ensure robustness across five historical editions. We also extend the KuiSCIMA dataset to include all 109 pieces from Baishidaoren Gequ, covering *suzipu*, *lülüpu*, and *jianzipu* notations.

Our contributions can be summarized as:

- Improved OMR models for *suzipu* and *lülüpu* with significantly lower CERs.
- Well-calibrated models with an Expected Calibration Error on 10 equidistant bins ($ECE_{10}$) below 0.0162.
- A comprehensive dataset extension, enabling future research on historical Chinese music.
- A comparison with state-of-the-art Chinese OCR and human performance, demonstrating the superiority of our models.

These contributions make the works of Jiang Kui accessible to both researchers and the public. With cultural artifacts in digital form, a great step towards the preservation and understanding of cultural heritage is achieved.

## 2   Related Works

This paper is interdisciplinary in the sense that it intersects the realms of datasets and document recognition of historical Chinese texts (Group 1), the creation of machine-readable symbolic ground truth music datasets (Group 2), and finally it is concerned with the development and evaluation of OMR methods (Group 3). For each of those three groups, some examples are given to position our work relative to them.

Group 1 contains databases such as the MTHv2 dataset in [13], with 1081663 character instances on 3199 pages in total. In [19], the ICDAR 2019 HDRC CHINESE dataset is introduced with textline annotations on 12850 pages. The SCUT-CAB database [7] contains detailed layout annotations of 4000 pages. The Kanseki Repository [23] and the Chinese Text Project [21] contain binarized images of pages and their texts per page, also containing editions of *Baishidaoren Gequ*, but the musical notations are not transcribed. Here, we see the potential of cross-referencing these databases with KuiSCIMA to include as much symbolic information as possible.

In Group 2, we can find OMR datasets such as MUSCIMA++ [8] for handwritten common practice period staff notation, the CAPITAN dataset [5] of mensural notation, or the GregoBase dataset [3] for Gregorian chants. However, there is a severe lack of ground-truth OMR datasets of East Asian music scores, and corpora such as the *Jingju* 京剧 dataset from the CompMusic project [16] contain annotated copyrighted scores, limiting the dissemination and use of the dataset for OMR purposes.

Regarding Group 3, we focus on OMR of East Asian music traditions, which is an under-researched topic. Works include Chen & Sheu [6], where different algorithms for recognition of *gongchepu* 工尺谱 scores of *Kunqu* 昆曲 musical theater are discussed. Shen et al. [20] focused on autonomous extraction methods of *jianzipu* notation. Unfortunately, these papers do not provide publicly available datasets that could be used for comparison against new OMR algorithms.

One very notable example belonging to both Group 2 and Group 3 is [9] by Han et al., in which a complete pipeline starting from digital encoding, creation of a machine-readable dataset, optical music recognition and stylistic composition of Korean court music using the *jeongganbo* 정간보 notation is presented. Among OMR approaches of East Asian music traditions, this paper is also exceptional in that it provides a publicly available dataset. One of the main goals of our work is to make the involved source code and datasets public, according to the principles of Open Science, striving for reproducible and transparent outcomes.

## 3 Preliminaries

All of the editions of *Baishidaoren Gequ* preserved today are copies from other copies. Even the earliest editions present today stem from the $18^{th}$ century and can be traced back to the document *Tao MS*, which is itself a direct copy of *Baishidaoren Gequ*, but also went missing [24]. Since the *suzipu* notation fell out of use much before that, the copies are unreliable as they were made by people not proficient in the notation's tradition.

Each *suzipu* symbol is made up of a pitch (11 classes) and an optional secondary component. Since the collection is the largest historical source of *suzipu* notation [22], containing only 17 pieces, even domain experts do not agree about the number or semantics of these secondary symbols. In this paper, the system established by Wu Santu in [25] is used, using 7 classes (6 proper classes or their absence), therefore yielding 77 classes in total. However, only 66 of those classes actually appear in the KuiSCIMA dataset, and their distribution is highly imbalanced (see Figure 6). Also, the appearance of symbols in some classes is often ambiguous and may vary considerably [17].

Table 1: Overview of the *suzipu* pitch (left), secondary (middle), and *lülüpu* (right) classes. The pitch symbols are derived from cursive writing of their names.

| *Suzipu* (Pitch) Symbol | Name | Pinyin | *Suzipu* (Secondary) Symbol | Name | Pinyin | *Lülüpu* Name | Pinyin |
|---|---|---|---|---|---|---|---|
| ム | 合 | *He* |  | (None) |  | 黄钟 | *Huangzhong* |
| マ | 四 | *Si* | め | 大顿 | *Dadun* | 大吕 | *Dalü* |
| 一 | 一 | *Yi* | リ | 小住 | *Xiaozhu* | 太簇 | *Taicu* |
| 幺 | 上 | *Shang* | フ | 丁住 | *Dingzhu* | 夹钟 | *Jiazhong* |
| レ | 勾 | *Gou* | カ | 大住 | *Dazhu* | 姑洗 | *Guxian* |
| 人 | 尺 | *Che* | ら | 折 | *Zhe* | 仲吕 | *Zhonglü* |
| フ | 工 | *Gong* | J | 拽 | *Ye* | 蕤宾 | *Ruibin* |
| リ | 凡 | *Fan* |  |  |  | 林钟 | *Linzhong* |
| ス | 六 | *Liu* |  |  |  | 夷则 | *Yize* |
| め | 五 | *Wu* |  |  |  | 南吕 | *Nanlü* |
| る | 高五 | *Gao Wu* |  |  |  | 无射 | *Wuyi* |
|  |  |  |  |  |  | 应钟 | *Yingzhong* |
|  |  |  |  |  |  | 黄钟清 | *Huangzhong Qing* |
|  |  |  |  |  |  | 大吕清 | *Dalü Qing* |
|  |  |  |  |  |  | 太簇清 | *Taicu Qing* |
|  |  |  |  |  |  | 夹钟清 | *Jiazhong Qing* |
|  |  |  |  |  |  | 折字 | *Zhezi* |

In comparison with *suzipu*, the *lülüpu* notation is simpler, consisting of 17 classes, all occurring in the dataset. This notation consist of a fixed subset of standard Chinese characters with a restricted number of classes. The *suzipu* and *lülüpu* symbols are shown in Table 1.

In the most complex notation, *jianzipu*, symbols consist of multiple components that are combined, making the class number very high[3]. The 708 dataset instances are made up of 98 different annotations. Due to the complexity and the lack of datasets, OMR of this notation cannot be performed in the setting of the KuiSCIMA dataset and is not considered in this paper.

Finally, the presence of musical or poetic semantics (such as musical modes or relationships between lyrics and melody) distinguishes the tasks of OCR and OMR. Together with the scarcity of data in total, few annotated data, and the aforementioned ambiguity the OMR tasks addressed in this paper are challenging.

---

[3] E.g., even when restricting ourselves to the simplest *anyin* 按音 plucks, i.e., a single right hand finger plucks a string while it is stopped with a finger of the left hand at an integer position, we get $8 \cdot 7 \cdot 4 \cdot 14 = 3136$ different combinations.

## 4  Experiments

Our experiments concern the task of symbol classification and clustering for both *suzipu* and *lülüpu* notation symbols taking image patches as input. The patches are extracted and stored as individual images with a JSON file describing their annotations. The source code is open-source and publicly available.[4]

The experiments compare the performance of proposed neural network based classification methods with classification by humans for *suzipu*, and with state-of-the-art Chinese character classifiers for *lülüpu*.

### 4.1  Model Architecture and Evaluation Methodology

Given the reduced size of both datasets (*suzipu* and *lülüpu*), the chosen CNN architecture is small to prevent overfitting. This has the additional benefit of reduced inference times which allow the model to run on the CPU.

The model comprises three convolutional layers and two fully connected layers, processing $48 \times 48$ pixel inputs. Each convolutional layer uses padding to maintain output dimensions, followed by ReLU activation, 2D batch normalization, and $2\times2$ max-pooling for dimensionality reduction. The first fully connected layer (`fc1`) includes ReLU activation and 50% dropout. The output dimension varies based on the feature extractor: 11 for *suzipu* pitch, 7 for *suzipu* secondary, or 17 for *lülüpu*. The architecture is illustrated in 1.

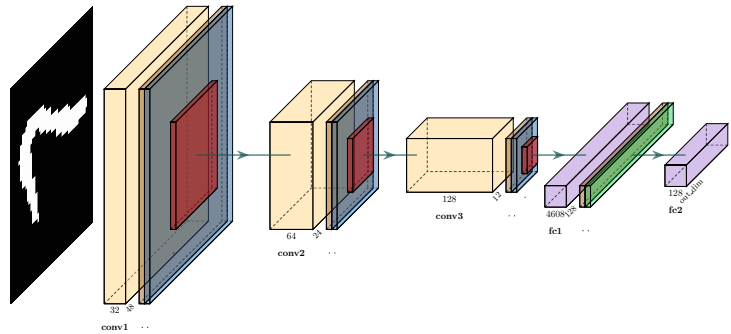

Fig. 1: A compact CNN with three convolutional and two fully connected layers.[5]

Each edition is handwritten, either manuscript or xylographic, with distinct variations in individual instances. Our evaluation is formulated as a leave-one-out cross validation: for each of the five editions, a model is trained to classify notation symbols of an unseen edition, the test set. The training and validation sets are formed by dividing samples according to their annotation (consisting of both pitch and secondary labels), ensuring 75% of samples of each class are randomly assigned to the training set and 25% to the validation set. Hereby, the composition of training and validation sets are guaranteed to be similar. For each cross-validation group and feature type (i.e., either pitch or secondary features) 10 distinct models are trained . In total, this yields 100 model samples for *suzipu*, and 50 for *lülüpu*. We refer to the model groups by the edition that constitutes their test set: `Lu`, `Zhang`, `Siku`, `Zhu`, and `Shanghai` models are tested on the edition of their name and trained with samples of every other edition.

---

[4] The link will be published after paper acceptance.
[5] The visualization was created using Haris Iqbal's PlotNeuralNet (https://github.com/HarisIqbal88/PlotNeuralNet).

### 4.2   Classifier for *Suzipu*

The original KuiSCIMA dataset contains 7297 non-empty annotated *Suzipu* instances, of which 133 are excluded from training due to anomalous shapes. Of the result, 5168 are simple symbols and 1996 are composite symbols.

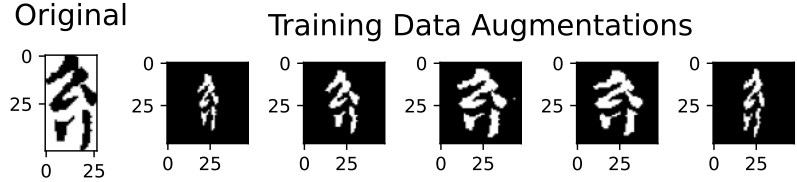

Fig. 2: These data augmentations are used for training the model.

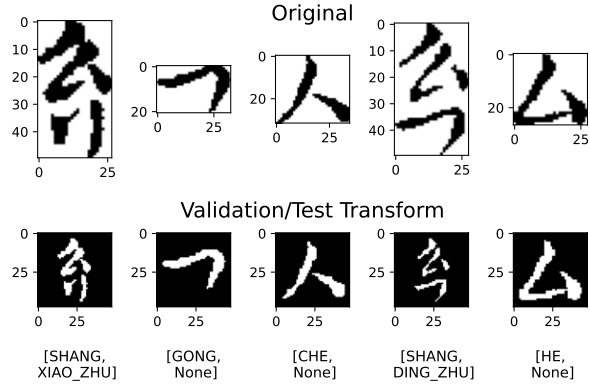

Fig. 3: Here, the transformation of the validation and test data is visualized.

**Image Preprocessing.** The preprocessing pipeline follows the original KuiSCIMA paper [17]. Training images are resized such that the longest side is uniformly scaled between 30 and 42 pixels, randomly rotated between -9 and 9 degrees, and cropped to $48 \times 48$ squares (see Figure 2). For validation and test images, the longest side is resized to 40 pixels, padded to a square, and normalized as described (see Figure 3).

**Model Training.** Separate instances of the model are used to predict pitch (11 output parameters) and secondary property (7 output parameters) resulting in a classifier made of two small CNNs leveraging the structure of the *suzipu* notation. For each component and test edition, 10 models are trained. Similar approaches have succeeded in settings where data can be seen as a product space, such as Ethiopic script [1].

Each model is trained with 80 epochs of 43 batches with 100 images each, sampling the training instances uniformly (with replacement) with respect to the feature class the classifier is trained on. The validation and test instances are traversed completely and without replacement.

The optimizer *Adam* with $10^{-3}$ learning rate is used, and a weight decay of $10^{-4}$ on all of the model's parameters except biases and batch normalization layers. In case the validation loss plateaus for five epochs, the learning rate is reduced by a factor of 0.5. The loss function is focal loss [12] by Lin et al. with $\gamma = 1$, since it reduces the loss of well-classified examples and therefore allows the model to learn from a small number of hard samples effectively (beneficial in settings of class imbalance).

**Results.** The average validation and test accuracies are shown in Table 2-top for the individual pitch and secondary classifiers (10 samples each per model class), as well as the joint pitch-secondary classifiers (100 samples per model class). The aggregated total CER is $6.6\% \pm 1.2\%$. The models are well-calibrated with an $ECE_{10}$ below 0.0154.

For each classifier and test set, the best performing model on the validation set is evaluated on the test set. Results (Table 2-bottom) reveal a total character error rate (CER) for the *suzipu* OMR task between 4.6% and 9.0%. `Zhu` achieves the highest CER for the averaged values. Even when selecting the best model according to validation accuracy, its outcomes are worse than the average for this edition. This indicates that the Zhu edition has unique features that cannot be properly approximated by the validation set.

For timing, an instance of the joint pitch-secondary classifiers was evaluated 500 times on the Shanghai MS (1439 instances). The tests were conducted using an AMD Ryzen 7 PRO 4750U processor. This yielded an average time of 2.04 $\pm$ 0.22 s for a single evaluation of the whole edition.

### 4.3   Classifier for *Lülüpu*

The *lülüpu* notation dataset used for training was introduced in KuiSCIMA v2.0 [17] and contains 3385 non-empty *lülüpu* instances.

Table 2: For the *suzipu* classifiers, the average and best validation/test accuracies and test CERs are displayed. The top table shows average accuracies with standard deviations, while the bottom table shows results for the best models (selected based on validation accuracy).

| Test Edition | Validation Acc. (%) | | Test Acc. (%) | | | Test CER (%) |
|---|---|---|---|---|---|---|
| | Pitch | Secondary | Pitch | Secondary | Total | Total |
| Lu | 97.0 ± 0.5 | 97.2 ± 0.6 | 96.5 ± 0.4 | 97.8 ± 0.3 | 94.5 ± 0.4 | **5.5 ± 0.4** |
| Zhang | 97.3 ± 0.4 | 97.7 ± 0.3 | 96.5 ± 0.7 | 96.8 ± 0.3 | 94.3 ± 0.7 | **5.7 ± 0.7** |
| *Siku Quanshu* | 96.9 ± 0.2 | 97.4 ± 0.3 | 96.4 ± 0.4 | 96.8 ± 0.5 | 93.9 ± 0.6 | **6.1 ± 0.6** |
| Zhu | 97.1 ± 0.5 | 97.5 ± 0.5 | 94.4 ± 0.6 | 96.0 ± 0.4 | 91.6 ± 0.6 | **8.4 ± 0.6** |
| Shanghai MS | 97.0 ± 0.4 | 97.8 ± 0.5 | 95.8 ± 0.5 | 96.1 ± 0.5 | 92.6 ± 0.6 | **7.4 ± 0.6** |
| Aggregated | | | 95.9 ± 1.0 | 96.7 ± 0.7 | 93.4 ± 1.2 | **6.6 ± 1.2** |

| Test Edition | Validation Acc. (%) | | Test Acc. (%) | | | Test CER (%) |
|---|---|---|---|---|---|---|
| | Pitch | Secondary | Pitch | Secondary | Total | Total |
| Lu | 97.7 | 98.2 | 97.0 | 97.3 | 94.6 | **5.4** |
| Zhang | 97.8 | 98.4 | 97.5 | 97.3 | 95.4 | **4.6** |
| *Siku Quanshu* | 97.3 | 97.9 | 96.5 | 97.7 | 94.7 | **5.3** |
| Zhu | 97.7 | 98.3 | 93.6 | 95.9 | 91.0 | **9.0** |
| Shanghai MS | 97.8 | 98.8 | 95.8 | 96.6 | 92.9 | **7.1** |

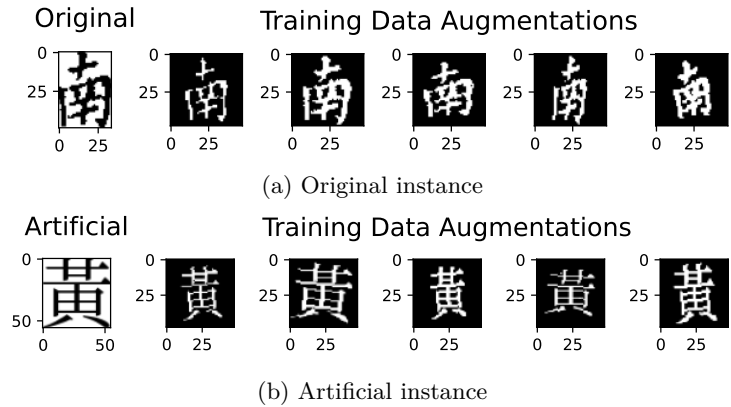

(a) Original instance

(b) Artificial instance

Fig. 4: This figure shows the training images with data augmentations.

**Image Preprocessing.** Due to the detailed structure of the notation, which uses Chinese characters for pitches, image patches do not undergo denoising. The preprocessing and augmentation pipeline follows the *suzipu* approach, with adjusted image dimensions to better capture fine details: training images are resized such that the longest side is uniformly scaled between 33 and 46 pixels, while test images are resized to 40 pixels, both resulting in a final size of $48 \times 48$ pixels.

To counter data scarcity, two classes of models are trained: One with original data only, and another one with artificially generated training data. For the latter, four different computer fonts (*AR PL UKai CN*, *AR PL Mingti2L Big5*, *Noto Sans CJK JP*, *Noto Serif CJK JP*), are used to generate all 17 classes with a font size of 60, then preprocessed as the original images. These fonts are chosen to expose the model to a variety of Chinese character styles, thus reducing overfitting. When supported by the fonts, the characters *huang* 黄 and *qing* 清 are added with their variant characters *huang* 黃 and *qing* 清. This yields additional 86 training images. Figure 4 shows training data and augmentations.

**Model Training.** The output dimension of the model is set to 17 matching the 17 classes in *lülüpu*. Each model is trained using 50 epochs of 21 batches (22 for artificial data) with 100 images each, with or without artificial data. As in the case of *suzipu*, the training instances are sampled uniformly (with respect to their annotation) with replacement. Validation and test instances are sampled without replacement. *Adam* optimizer with LR of $5 \cdot 10^{-4}$ is used. Weight decay, learning rate scheduling and loss function are the same as for *suzipu*.

**Results.** Table 3-top shows the average validation and test accuracies for each model with and without artificial data (10 samples each per model class).

For each classifier and test set, the best performing model on the validation set is evaluated on the test set. Table 3-bottom reveals the aggregated CER for the *lülüpu* OMR task is 1.7% for models without artificial data and 0.9% when including the artificial data. Also, the introduction of artificial samples reduced the variance of the accuracy values, making these models more robust. Using temperature scaling, we lowered the maximum $ECE_{10}$ from 0.0177 to 0.0162.

As for the *suzipu* classifier, Zhu achieved the worst results with the best model according to validation accuracy performing worse than the average accuracy. However, the mean accuracy and the variance for aggregated CER values improved notably when using artificial data.

Regarding timing, an instance of the model evaluated 500 times yields an average time of $0.53 \pm 0.07$s for a single evaluation of the whole Shanghai MS (which contains 677 *lülüpu* instances).

**Comparison with Tesseract OCR.** Since the *lülüpu* notation uses a subset of Chinese characters, a comparison of the trained model against an out-of-the-box OCR algorithm is meaningful.

Table 3: For the *lülüpu* classifiers, the top table shows average validation/test accuracies with standard deviations, while the bottom table displays results for the best models (selected based on validation accuracy). The best test CER values are highlighted in bold.

| Test Edition | Validation Acc. (%) | | Test Acc. (%) | | Test CER (%) | |
| --- | --- | --- | --- | --- | --- | --- |
| | Non-Art. | Art. | Non-Art. | Art. | Non-Art. | Art. |
| Lu | 99.4 ± 0.2 | 99.6 ± 0.2 | 99.5 ± 0.2 | 99.7 ± 0.2 | 0.5 ± 0.2 | **0.3 ± 0.2** |
| Zhang | 99.4 ± 0.2 | 99.5 ± 0.2 | 99.4 ± 0.2 | 99.7 ± 0.2 | 0.6 ± 0.2 | **0.3 ± 0.2** |
| *Siku Quanshu* | 99.4 ± 0.3 | 99.4 ± 0.2 | 99.3 ± 0.4 | 99.5 ± 0.3 | 0.7 ± 0.4 | **0.5 ± 0.3** |
| Zhu | 99.6 ± 0.3 | 99.6 ± 0.3 | 95.1 ± 1.9 | 97.8 ± 1.3 | 4.9 ± 1.9 | **2.2 ± 1.3** |
| Shanghai MS | 99.6 ± 0.2 | 99.6 ± 0.2 | 98.2 ± 0.6 | 98.7 ± 0.4 | 1.8 ± 0.6 | **1.3 ± 0.4** |
| Aggregated | | | 98.3 ± 1.9 | 99.1 ± 0.9 | 1.7 ± 1.9 | **0.9 ± 0.9** |

| Test Edition | Validation Acc. (%) | | Test Acc. (%) | | Test CER (%) | |
| --- | --- | --- | --- | --- | --- | --- |
| | Non-Art. | Art. | Non-Art. | Art. | Non-Art. | Art. |
| Lu | 99.7 | 99.9 | 99.7 | 99.6 | **0.3** | 0.4 |
| Zhang | 99.7 | 99.7 | 99.4 | 99.7 | 0.6 | **0.3** |
| *Siku Quanshu* | 99.7 | 99.9 | 99.3 | 99.6 | 0.7 | **0.4** |
| Zhu | 100.0 | 99.9 | 95.0 | 95.4 | 5.0 | **4.6** |
| Shanghai MS | 100.0 | 99.9 | 97.6 | 99.1 | 2.4 | **0.9** |

Images are preprocessed by resizing to 60×60 pixels (cubic interpolation), applying a 3×3 Gaussian blur, binarizing with Otsu's method, and padding with 20-pixel whitespace on all sides. The *Tesseract* OCR system [10] (using the model `chi_tra.traineddata`[6]) is applied with a whitelist of 15 *lülüpu* characters and three page segmentation modes (PSM 6, 7, 8). Spaces are removed from the OCR output, and predictions are matched to *lülüpu* classes. If the output contains *zhe* 折 or *zi* 字, the *Zhezi* 折字 class is returned. If no match is found, the most frequent class, *Linzhong* 林钟 (11.9% of the dataset), is assigned. The test is performed on the entire KuiSCIMA *lülüpu* dataset.

The CER is for each PSM 6 = 42.4%, for PSM 7 = 42.1%, and for PSM 8 = 44.0%. We assume three reasons for these results: First, Tesseract models, though whitelisted, are optimized for a large character set, reducing performance on small subsets. Second, they are likely trained on modern fonts, limiting effectiveness on handwritten historical texts. Third, OCR typically relies on contextual information, which is unavailable for isolated notation characters in OMR. The timing test was performed with 15 iterations on the Shanghai MS dataset, yielding an average time of 472.2 ± 4.8 seconds.

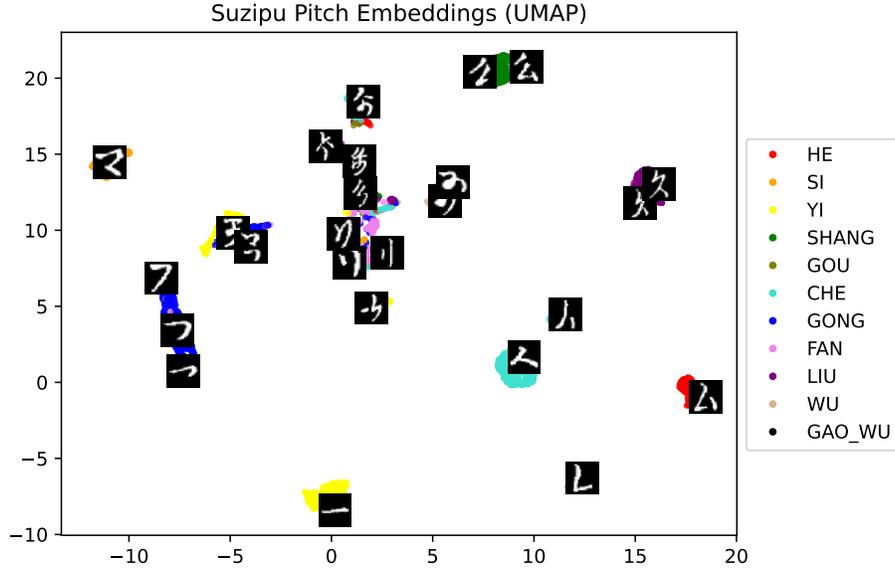

Fig. 5: *Suzipu* pitch classifier UMAP embedding. Some samples are drawn at their corresponding positions.

### 4.4 Similarity Visualization

Similarity visualization displays for the annotation tool are elaborated in this section. For a selected instance, the three most optically similar notation instances found in in the KuiSCIMA v2.0 dataset.

To obtain sharp cluster separations, the five best *suzipu* pitch and secondary models for each test edition (Section 4.2), and the five best *lülüpu* models (Section 4.3) are applied to each individual sample of the KuiSCIMA dataset. Selecting the best performing model results in the `Zhang` model (98.6% accuracy) for the *suzipu* pitch classifier, and the `Lu` model for both the *suzipu* secondary (98.9% accuracy) and *lülüpu* classifiers (99.82% accuracy).

For each of the three resulting OMR models, the layer `fc1` (which is a 128-dimensional vector) is extracted as a feature encoding of the respective property. The features are collected for each instance in the dataset, and an unsupervised UMAP [15] dimensionality reduction to 2D is applied with a `random_state` of 42. Figures 5 illustrates an exemplary visualization of the *suzipu* pitch UMAP spaces including samples from the dataset (color matching ground truth label).

The UMAP representation of the currently investigated notation instance is compared against the precomputed UMAP representation of all instances of the same notation type in the KuiSCIMA dataset, and the three nearest neighbors are retrieved and displayed in the user interface.

---

[6] https://github.com/tesseract-ocr/tessdata/blob/main/chi_tra.traineddata

## 5 *Suzipu* Classification Human-Level Performance

To compare performance (accuracy, time) of our *suzipu* classifier, a user study was conducted with minimally trained and naive participants (with respect *suzipu* notation) perform the task of *suzipu* classification.

Similar to Belay's study [2] for the Ethiopic script[7], the human annotations are collected by users at home using a specialized Web GUI. The source code for the software used in this study is open-source and publicly available[8].

### 5.1 Participants.

A sample of 15 participants (7 female, 8 male; aged $33 \pm 9.74$ years) were recruited as volunteers through experiment postings. The participants have varying degrees of formal education. All of them have normal or corrected visual acuity. Participants were naive to *suzipu* notation and Chinese language, corresponding to settings as in Belay's [2], where in highly specialized tasks, a large number of minimally trained human annotators collect a dataset which is then corrected by a small number of highly skilled experts.

### 5.2 Task description.

The study consists of a set of tasks, where each task corresponds to an image of *suzipu* notation. The goal is to identify the correct pitch and secondary components in the *suzipu* framework as described in Section 3.

All of the tasks are taken from the Shanghai MS, the earliest edition preserved until today, with the shape of the symbols being clearest among all editions. This edition's *suzipu* notations were partitioned into two subsets according to their composite annotations (i.e., pitch and secondary), such that each of the two subsets have a similar distribution of samples according to their labels with 719 respectively 720 samples.

Each participant receives a sequential number according to the order of participation, corresponding to the dataset partition they are given to annotate.

### 5.3 Study design.

The study unfolds as follows: A training phase allows the user to familiarize with the tasks. A written task description, a demo video of five minutes, and a training software (providing the user 10 tasks representative of the measurement phase).

After training, participants independently complete the measurement phase on the study website. They annotate samples from their corresponding dataset partition, with each action timestamped. Accuracy metrics (pitch, secondary, and joint pitch-secondary) and completion time are recorded, with time calculated as the sum of action intervals (capped at 5 min per interval). Users were instructed to perform tasks uninterrupted in a calm environment to ensure comparability and traceability.

---

[7] https://github.com/bdu-birhanu/HHD-Ethiopic/blob/main/Supplementary_file/Revised_Historical_Handwritten_Ethiopic_dataset_Sup.pdf

[8] The link will be published after paper acceptance.

The user study has been approved by the Ethics Committee of *Name will be disclosed after paper acceptance* (reference number *00000000*).

### 5.4  Study results.

Table 4: Accuracy values pitch/secondary/joint classification, average/best user.

|  | Accuracy (%) | | | CER (%) |
|---|---|---|---|---|
|  | Pitch | Secondary | Total | Total |
| Average | $87.2 \pm 10.0$ | $90.8 \pm 5.0$ | $84.1 \pm 10.0$ | $15.9 \pm 10.0$ |
| Best | 94.7 | 96.1 | 92.4 | 7.6 |

All 15 participants completed the study successfully. The average time needed to complete the study was $76.3 \pm 19.3$ minutes. Accounting for the different numbers of presented samples, the average time to complete a whole edition is therefore $152.5 \pm 38.6$ minutes. The aggregated metrics (pitch accuracy, secondary accuracy, total accuracy, and total CER) and the metrics of the best performing participant are shown in Table 4. Results exhibit high standard deviation. The best user's CER is 7.6% . Comparing the users' accuracies with the OMR models' using a Wilcoxon rank-sum test ($\alpha = 0.01$) yields that the median OMR accuracy is significantly ($p = 1.9 \cdot 10^{-7} < \alpha$) larger than the humans'. This non-parametric test was chosen due to the small sample size and the stark negative skew in the human accuracy values. In addition, the OMR classifiers' worst 25%-quantile is contained inside the humans' best 25%-quantile.

## 6  Interpretation

The *suzipu* classification task, involving $11 \cdot 7 = 77$ classes, resulted in an aggregated CER of $6.6\% \pm 1.2\%$. In comparison, regarding the similar task of Ethiopic script character classification (with a decomposition into $33 \cdot 7 = 231$ classes), the model in [1] obtained a CER of 5.03%, with a dataset consisting of 77994 instances (ca. ten times larger than *suzipu* in the KuiSCIMA dataset).

Results on the Shanghai MS (CER = 7.1%) reveal a clear improvement against the previous baseline model of the original KuiSCIMA dataset in [17] (CER = 10.4%). This value is not only better than the average human-level performance (CER = $15.9\% \pm 10.0\%$), it also outperforms the best-performing human user (CER = 7.6%). This strongly indicates that the OMR models are reliable for the task of *suzipu* classification, despite scarcity of the training dataset.

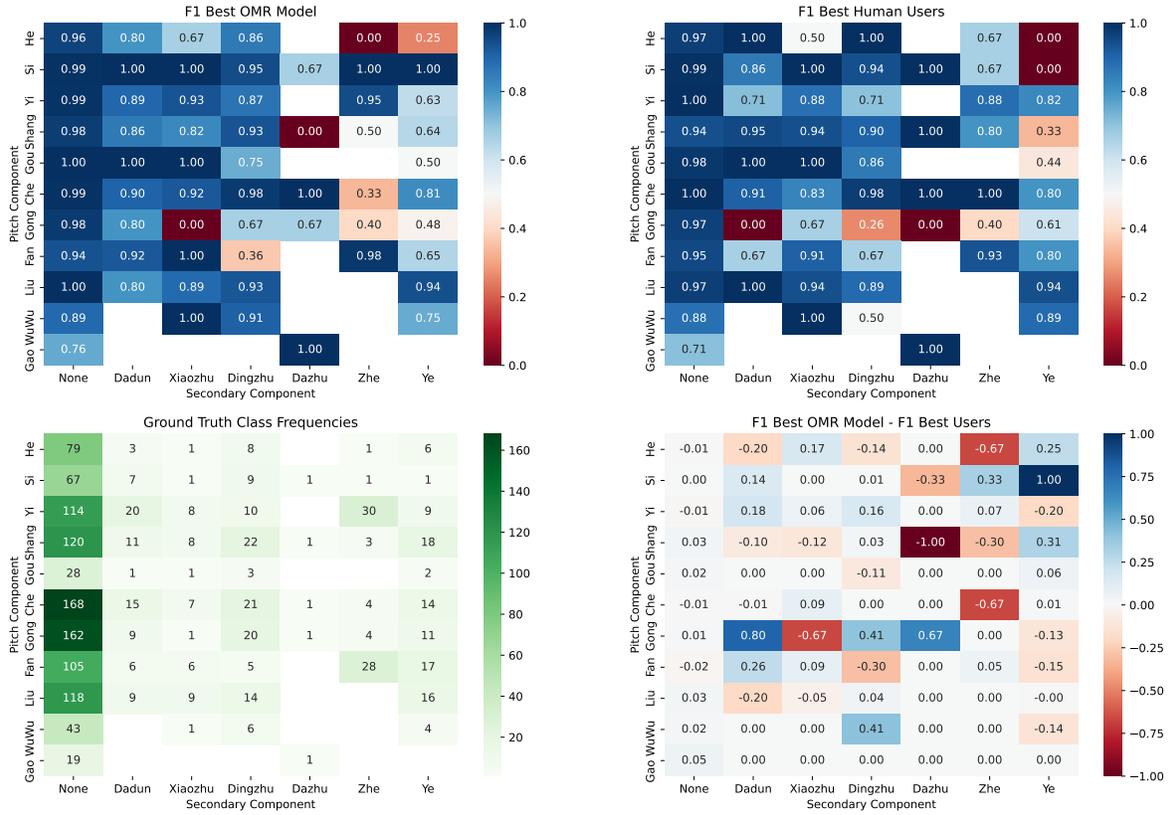

Fig. 6: The heatmaps show the F1 scores for the best OMR model (top left), the best two human users (top right), their difference (bottom right) and the absolute class frequency in the Shanghai MS (bottom left).

The per-class F1 scores for the best OMR model and the two best human users (spanning the whole Shanghai MS) in Figure 6 indicate that both the model and the humans are similar in that they both perform well on very frequent classes (e.g., the class *Che/None*). Vice versa, classes where humans (e.g., *Gong/Dadun*) or the model (e.g., *Gong/Xiaozhu*) perform badly are infrequent classes. The average per-class F1 scores for the model ($0.79 \pm 0.26$) and the human user ($0.79 \pm 0.27$) exhibit a similar performance on individual class level, even though the individual classes where F1 is low are different.

Regarding time needed, the *suzipu* model needed $2.04 \pm 0.22$ seconds, while it took the average user $9152 \pm 2318$ seconds. Thus, the classifier's accuracy is higher and the time needed is significantly lower than the human users'.

The *lülüpu* classification task involves 17 distinct classes, and the OMR models perform with an aggregated CER= $0.9\% \pm 0.9\%$. The tesseract CER model's performance is not comparable (CER = $51.6\%$). Also regarding timing, the *lülüpu* CNN is with $0.53 \pm 0.07$ seconds (500 evalutions) significantly faster than the Tesseract model with $472.2 \pm 4.8$ seconds.

The test results on both the *suzipu* and *lülüpu* classifiers exhibit significantly lower accuracies when the Zhu edition is taken as a test set. It is an indication that the distribution of the Zhu dataset's image data differs from that of the other editions. Considering prior results on variability in [14], one can conclude that all other editions do not introduce as much data variability each as the Zhu edition, since for them, high accuracy values are achieved despite dataset exclusion in training and vice versa.

The classifiers are well-calibrated with ECE values (for 10 equidistant bins) in the range of 0.0062 to 0.0162, thus constituting well-calibrated models. Using UMAP, the classifiers' feature embeddings have been visualized, and the distinct separation between clusters corresponds well to ground-truth labels.

## 7    Conclusion and Future Work

With the introduction of the KuiSCIMA v2.0 dataset, an important step towards dissemination and preservation of cultural heritage, and the foundation for computational studies of the involved notations has been established. Since all of our contributions are open source and publicly available[9], a great impact on OMR research of culturally diverse music traditions is expected.

We developed well-calibrated and reliable OMR algorithms for the *suzipu* and *lülüpu* notations. In the case of *suzipu*, an annotator must only correct around every 15th model prediction instead of manually annotating every single instance. Equivalently, for *lülüpu*, an expert is required to rectify only every 100th prediction. These methods have been integrated into a specialized annotation GUI [18], achieving a significant reduction of human effort.

In the future, we propose new research based on KuiSCIMA as follows:

---

[9] The link to the dataset will be provided after acceptance of the paper.

- Re-encoding the digital representations in KuiSCIMA into a standard notation format like the Music Encoding Initiative (MEI[10]).
- Digitizing other *suzipu*-based music, e.g., *Xi'an Guyue* 西安鼓乐, and applying OMR methods to similar music traditions such as *Jyutkek* 粵劇.
- Developing end-to-end OMR systems for *suzipu* and *lülüpu* notations.

**Disclosure of Interests.** The authors have no competing interests to declare that are relevant to the content of this article.

---

[10] http://www.music-encoding.org